\documentclass[10pt,twocolumn,letterpaper]{article}

\usepackage[accepted]{cvpr}

\definecolor{cvprblue}{rgb}{0.21,0.49,0.74}
\usepackage[pagebackref,breaklinks,colorlinks,allcolors=cvprblue]{hyperref}

\def\confName{CVPR}
\def\confYear{2026}

\usepackage{multirow}
\usepackage{microtype}
\usepackage{colortbl}

\title{Consistent but Dangerous: Per-Sample Safety Classification Reveals False Reliability in Medical Vision-Language Models}

\author{Binesh Sadanandan\\
SAIL Lab, University of New Haven\\
West Haven, CT, USA\\
{\tt\small bsada1@unh.newhaven.edu}
\and
Vahid Behzadan\\
SAIL Lab, University of New Haven\\
West Haven, CT, USA\\
{\tt\small vbehzadan@newhaven.edu}}

\begin{document}
\maketitle

\begin{abstract}
Consistency under paraphrase, the property that semantically equivalent prompts yield identical predictions, is increasingly used as a proxy for reliability when deploying medical vision-language models (VLMs).
We show this proxy is fundamentally flawed: a model can achieve perfect consistency by relying on text patterns rather than the input image.
We introduce a \emph{four-quadrant} per-sample safety taxonomy that jointly evaluates \textbf{consistency} (stable predictions across paraphrased prompts) and \textbf{image reliance} (predictions that change when the image is removed).
Samples are classified as \emph{Ideal} (consistent and image-reliant), \emph{Fragile} (inconsistent but image-reliant), \emph{Dangerous} (consistent but not image-reliant), or \emph{Worst} (inconsistent and not image-reliant).
Evaluating five medical VLM configurations across two chest X-ray datasets (MIMIC-CXR, PadChest), we find that LoRA fine-tuning dramatically reduces flip rates but shifts a majority of samples into the Dangerous quadrant: LLaVA-Rad Base achieves a 1.5\% flip rate on PadChest while 98.5\% of its samples are Dangerous.
Critically, Dangerous samples exhibit \emph{high accuracy} (up to 99.6\%) and \emph{low entropy}, making them invisible to standard confidence-based screening.
We observe a negative correlation between flip rate and Dangerous fraction ($r=-0.89$, $n{=}10$) and recommend that deployment evaluations always pair consistency checks with a text-only baseline: a single additional forward pass that exposes the false reliability trap.
\end{abstract}

\section{Introduction}
\label{sec:intro}

Medical vision-language models (VLMs) are rapidly approaching clinical deployment for tasks such as visual question answering on chest radiographs~\cite{medgemma2025,chaves2024llavarad,llavamed2024}.
A critical requirement for such systems is \emph{reliability}: clinicians must trust that equivalent clinical queries produce equivalent answers.
This has motivated a focus on \emph{paraphrase consistency}, the invariance of model predictions under semantically equivalent reformulations of the input question~\cite{elazar2021measuring,ribeiro2020beyond}.

The dominant approach measures consistency through the \emph{flip rate}: the fraction of samples for which at least one paraphrase elicits a different prediction than the original question~\cite{sadanandan2026psfmed}.
A low flip rate is interpreted as evidence of stable, reliable behavior.
We argue that this interpretation contains a critical blind spot.

Consider a radiologist assistant that, when asked ``Is there a pleural effusion?'' or ``Can pleural effusion be identified?'', consistently answers ``yes'' regardless of the chest radiograph shown.
Such a model achieves perfect consistency (0\% flip rate) and may even achieve reasonable accuracy on imbalanced datasets where effusions are common.
Yet it is manifestly unsafe: it produces the appearance of reliability while producing the same prediction when the image is removed under our test.
A clinician evaluating this system using flip rate and accuracy alone would conclude it is ready for deployment. That conclusion could lead to missed diagnoses or inappropriate treatment.

This is not a hypothetical failure mode.
We demonstrate that it characterizes the majority behavior of several medical VLMs, including fine-tuned models specifically optimized for paraphrase consistency.
Across five model configurations and two chest X-ray datasets, we find that the models with the \emph{lowest} flip rates, those that would pass consistency-based deployment checks, have the \emph{highest} fraction of predictions whose output is unchanged when the image is removed.

\paragraph{Contributions.}
We make four contributions:
\begin{enumerate}[nosep,leftmargin=*]
\item We introduce a \textbf{four-quadrant per-sample safety taxonomy} that jointly evaluates consistency and image reliance, identifying the \emph{Dangerous} quadrant of consistent-but-text-reliant predictions (\cref{sec:framework}).
\item We evaluate \textbf{five medical VLM configurations} across two chest X-ray datasets, revealing a negative correlation ($r=-0.89$, $\rho=-0.79$, $n{=}10$) between flip rate and Dangerous fraction: models optimized for consistency have the highest fraction of text-reliant predictions (\cref{sec:results}).
\item We show that Dangerous samples exhibit \textbf{high accuracy and low entropy}, making them undetectable by standard uncertainty-based screening; they evade every standard safety check simultaneously (\cref{sec:entropy}).
\item We propose a practical \textbf{deployment recommendation}: always pair consistency evaluation with a text-only baseline, requiring only one additional forward pass per sample (\cref{sec:discussion}).
\end{enumerate}

\section{Related Work}
\label{sec:related}

\paragraph{Medical VLM evaluation.}
Medical VLMs have been evaluated on visual question answering~\cite{lau2018vqarad}, report generation~\cite{chaves2024llavarad}, and multi-task benchmarks~\cite{xia2024cares,zhang2024biomedgpt}.
Specialized models such as CheXagent~\cite{chen2024chexagent} and Med-Flamingo~\cite{moor2023medflamingo} have expanded the scope to chest X-ray interpretation and few-shot learning, while prompt engineering studies~\cite{nori2023gpt4medprompt} explore whether generalist models can match specialist tuning.
These evaluations primarily focus on aggregate accuracy, with limited attention to per-sample behavioral analysis.
Trustworthiness benchmarks~\cite{xia2024cares} assess fairness, privacy, and safety, and robustness evaluations~\cite{kahl2025surevqa} measure performance under perturbations, but none jointly assess consistency and image grounding at the sample level.

\paragraph{Paraphrase sensitivity.}
Consistency under paraphrase has been studied for language models~\cite{elazar2021measuring} and adapted to multimodal settings through behavioral testing frameworks~\cite{ribeiro2020beyond}.
The paraphrase sensitivity failure (PSF) framework~\cite{sadanandan2026psfmed} measures flip rates across controlled paraphrase phenomena including lexical substitution, syntactic restructuring, negation patterns, scope quantification, and specificity modulation.
However, existing work treats consistency as an unconditional positive: a model with zero flip rate is considered strictly better than one with nonzero flip rate.
We show that this ranking can be actively misleading, as the zero-flip-rate model may achieve consistency by relying on text patterns rather than visual input.

\paragraph{Image grounding and text shortcuts.}
Deep learning models are known to exploit spurious correlations and text shortcuts~\cite{geirhos2020shortcut}, and VLMs can hallucinate content not present in the image~\cite{li2023evaluating,huang2024hallucination}.
In medical VQA specifically, models can achieve high accuracy by exploiting statistical regularities in question text without examining the image~\cite{lau2018vqarad}; for example, questions about common findings (\eg, ``Is there cardiomegaly?'') may be answerable from base rates alone.
Our text-only baseline methodology operationalizes this insight: by comparing predictions with and without the image, we classify each sample's reliance on visual evidence rather than reporting only aggregate shortcut prevalence.

\paragraph{Uncertainty in medical AI.}
Predictive uncertainty has been proposed as a safeguard for clinical deployment~\cite{kompa2021second,guo2017calibration}, with the expectation that unreliable predictions will exhibit high entropy or low confidence.
Entropy-based methods~\cite{gal2016dropout,kadavath2022language} and semantic uncertainty~\cite{kuhn2023semantic} can flag predictions the model is unsure about.
We show that the Dangerous quadrant is precisely where uncertainty methods fail: these samples have high confidence and low entropy despite being not image-reliant under \cref{eq:image_reliant}.
This represents a qualitatively different failure mode from the calibration errors typically studied in the uncertainty literature~\cite{guo2017calibration}, because the model's confidence is well-calibrated \emph{for the wrong reason}: it is confident because it has learned a reliable text shortcut, not because it has correctly interpreted the image.

\section{Methods}
\label{sec:methods}

\subsection{Four-Quadrant Safety Framework}
\label{sec:framework}

We define two binary per-sample properties for a VLM on a given image-question pair:

\noindent\textbf{Consistency.}
Given a question $q$ and $K$ paraphrases $\{q_1, \ldots, q_K\}$ that are semantically equivalent to $q$, a sample is \emph{consistent} if the model prediction is identical for $q$ and all $q_k$:
\begin{equation}
\text{consistent}(x, q) = \mathbb{1}\left[\bigwedge_{k=1}^{K} f(x, q) = f(x, q_k)\right]
\label{eq:consistent}
\end{equation}
where $f(x, q)$ denotes the model's prediction for image $x$ and question $q$.
This is a strict definition: a single disagreement among $K$ paraphrases renders the sample inconsistent.
We adopt this strict criterion because clinical deployment requires reliability across \emph{all} reasonable phrasings, not just most.

\noindent\textbf{Image reliance.}
A sample is \emph{image-reliant} if the model's prediction changes when the image is removed:
\begin{equation}
\text{image\_reliant}(x, q) = \mathbb{1}\left[f(x, q) \neq f(\varnothing, q)\right]
\label{eq:image_reliant}
\end{equation}
where $f(\varnothing, q)$ is the text-only prediction with no image input.
This operationalization is intentionally binary and conservative: if the prediction is the same with and without the image, the model's consistency may be grounded in text patterns rather than visual evidence.
This provides an upper bound on text-reliant behavior; a model may use the image to calibrate confidence without changing the final prediction, which our binary test would not detect.

These two properties define four quadrants (\cref{fig:framework}):
\begin{itemize}[nosep,leftmargin=*]
\item \textbf{Ideal}: consistent $\wedge$ image-reliant. Stable predictions grounded in the image. The model gives the same answer regardless of phrasing, and that answer depends on visual input.
\item \textbf{Fragile}: $\neg$consistent $\wedge$ image-reliant. The model uses the image but is sensitive to prompt phrasing. These samples represent genuine paraphrase sensitivity failures.
\item \textbf{Dangerous}: consistent $\wedge$ $\neg$image-reliant. Stable predictions that would be identical without the image, a false appearance of reliability.
\item \textbf{Worst}: $\neg$consistent $\wedge$ $\neg$image-reliant. Unstable and ungrounded. Both unreliable and not using the image.
\end{itemize}

\begin{figure}[t]
\centering
\small
\setlength{\tabcolsep}{3pt}
\begin{tabular}{r|c|c|}
\multicolumn{1}{r}{} & \multicolumn{1}{c}{\textbf{Image-reliant}} & \multicolumn{1}{c}{\textbf{Not image-reliant}} \\
\cline{2-3}
\textbf{Consistent} & \cellcolor{green!20}Ideal & \cellcolor{red!20}Dangerous \\
\cline{2-3}
\textbf{Inconsistent} & \cellcolor{orange!20}Fragile & \cellcolor{purple!20}Worst \\
\cline{2-3}
\end{tabular}
\caption{Four-quadrant per-sample safety taxonomy. Consistency is necessary but not sufficient for safe deployment: the \emph{Dangerous} quadrant achieves consistency without image grounding.}
\label{fig:framework}
\end{figure}

The key insight is that the Dangerous quadrant is invisible to consistency-only evaluation.
A model with 100\% Dangerous samples reports 0\% flip rate, the best possible consistency score, while being decision-invariant to image removal under \cref{eq:image_reliant}.

\subsection{Models}
\label{sec:models}

We evaluate five VLM configurations spanning two model families and three adaptation strategies:

\noindent\textbf{MedGemma-4B-IT}~\cite{medgemma2025}: A 4-billion parameter medical VLM based on the Gemma~2 architecture with 34 transformer layers and 256 image tokens. We evaluate three configurations:
\begin{itemize}[nosep,leftmargin=*]
\item \emph{Base}: the pretrained checkpoint without adaptation.
\item \emph{Targeted LoRA}: low-rank adaptation~\cite{hu2022lora} applied to layers 15--19 (rank 16, $\alpha=32$, 0.1\% of parameters), trained on MIMIC-CXR train split to reduce paraphrase sensitivity.
Training minimizes symmetric KL divergence on yes/no logit distributions between original and paraphrased prompts, with the image always present.
\item \emph{Full LoRA}: adaptation applied to all 34 layers with the same objective and hyperparameters, representing the maximum-capacity fine-tuning configuration.
\end{itemize}

\noindent\textbf{LLaVA-Rad}~\cite{chaves2024llavarad}: A radiology-specialized VLM fine-tuned from LLaVA~\cite{liu2024llava} on radiology report data. We evaluate:
\begin{itemize}[nosep,leftmargin=*]
\item \emph{Base}: the published checkpoint.
\item \emph{LoRA}: fine-tuned for paraphrase consistency using the same training objective as the MedGemma variants.
\end{itemize}

\noindent\textbf{Text-only baselines.}
For each model, we obtain text-only predictions $f(\varnothing, q)$ by running a standard forward pass with the image omitted entirely: for MedGemma, the prompt is constructed without the \texttt{<start\_of\_image>} token and no image is passed to the processor; for LLaVA-Rad, the prompt omits the \texttt{<image>} placeholder and only text token IDs are provided.
In both cases, image tokens are absent (not replaced with zeros or noise), and yes/no logits are extracted from the last position using the same token IDs and procedure as the image-conditioned pass.
No generation, sampling, or temperature scaling is applied in either condition; both use a single deterministic forward pass.
No additional training or model modification is required.

\subsection{Datasets}
\label{sec:datasets}

\noindent\textbf{MIMIC-CXR}~\cite{johnson2019mimic}: Binary presence questions (``Is there [finding]?'') over frontal chest radiographs from the MIMIC-CXR test split, with 98 balanced test samples (approximately 50\% ``yes'' ground truth).
Each sample has $K=5$ paraphrases spanning controlled linguistic phenomena: lexical substitution, syntactic restructuring, negation patterns, scope quantification, and specificity modulation~\cite{sadanandan2026psfmed}.
The balanced label distribution means that a model answering randomly or always ``yes'' achieves approximately 50\% accuracy, providing a clear baseline.
LLaVA-Rad models yield 78 evaluable samples due to parsing differences in the evaluation pipeline.

\noindent\textbf{PadChest}~\cite{bustos2020padchest}: A flip bank of 861 samples drawn from a Spanish multi-label chest X-ray dataset, serving as an out-of-distribution (OOD) test set.
The ground truth is imbalanced (81\% ``yes''), making it particularly revealing for text-reliant behavior: a model that always answers ``yes'' achieves 81\% accuracy without consulting the image.
This imbalance is clinically realistic: most binary queries about specific findings yield positive answers in curated diagnostic datasets. It also exposes models that learn to predict the majority class.

\noindent\textbf{Data governance.}
MIMIC-CXR is accessed under the PhysioNet Credentialed Health Data License~1.5.0; PadChest under the BIMCV Data Use Agreement.
All images are de-identified at source; no images are redistributed.
The PSF-Med paraphrases are released under CC~BY~4.0~\cite{sadanandan2026psfmed}.

\subsection{Metrics}
\label{sec:metrics}

\noindent\textbf{Flip rate}: fraction of samples with at least one paraphrase disagreement:
\begin{equation}
\text{FR} = 1 - \frac{1}{N}\sum_{i=1}^{N} \text{consistent}(x_i, q_i)
\end{equation}

\noindent\textbf{Dangerous fraction}: fraction of consistent-but-not-image-reliant samples:
\begin{equation}
\text{DF} = \frac{1}{N}\sum_{i=1}^{N} \mathbb{1}[\text{consistent}(x_i, q_i) \wedge \neg\text{image\_reliant}(x_i, q_i)]
\end{equation}

\noindent\textbf{Per-quadrant accuracy}: accuracy computed separately within each quadrant, revealing whether high overall accuracy masks ungrounded predictions.

\noindent\textbf{Predictive entropy}: $H = -\sum_c p_c \log p_c$ where $p_c$ is the model's softmax probability for class $c$.
Low entropy indicates high confidence; we examine whether Dangerous samples can be detected through elevated entropy.

\noindent\textbf{Bootstrap confidence intervals}: 95\% CIs computed over $B=2000$ bootstrap resamples~\cite{efron1993bootstrap} for flip rate and Dangerous fraction to account for finite sample sizes.

\section{Results}
\label{sec:results}

\subsection{Quadrant Distributions}
\label{sec:distributions}

\begin{figure}[t]
\centering
\includegraphics[width=\linewidth]{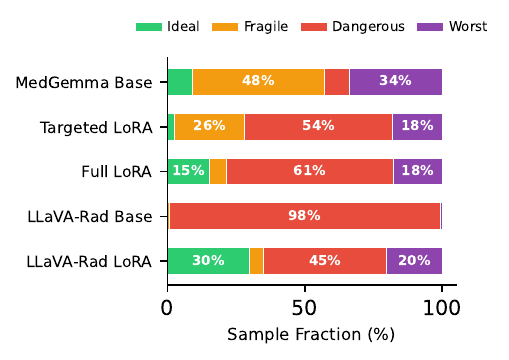}
\caption{Quadrant distribution on PadChest (861 samples for MedGemma variants, 732 for LLaVA-Rad). LLaVA-Rad Base is nearly entirely Dangerous (98.5\%), while MedGemma Base is primarily Fragile (47.9\%), meaning it uses the image but is paraphrase-sensitive.}
\label{fig:stacked_bar}
\end{figure}

\cref{tab:quadrant_dist} presents the full quadrant distribution across all model-dataset combinations, and \cref{fig:stacked_bar} visualizes the PadChest distributions.

The most striking finding is the prevalence of the Dangerous quadrant.
On PadChest, LLaVA-Rad Base places 98.5\% of samples in Dangerous: nearly every prediction is consistent but identical without the image, with zero Ideal samples.
Full LoRA places 60.7--78.6\% in Dangerous across both datasets; Targeted LoRA 53.8--58.2\%.
Even the best model on this metric (MedGemma Base, 9.1\% Dangerous on PadChest) achieves this only at the cost of the highest flip rate (81.7\%).
The distribution shift between MIMIC (balanced, 50\% ``yes'') and PadChest (imbalanced, 81\% ``yes'', OOD) amplifies the paradox: PadChest's label prior makes the ``always yes'' shortcut both viable and accurate, inflating Dangerous fractions up to 98.5\%.
Conditioning on ground-truth label, the Dangerous rate on PadChest exists for both ``yes'' and ``no'' samples (Full LoRA: 100\% of GT-``no'' vs.\ 52\% of GT-``yes''; Targeted LoRA: 76\% vs.\ 49\%), confirming that the effect is not solely an artifact of majority-class prediction.
On MIMIC, Dangerous fractions are lower (25.5--78.6\%) but arise from a more concerning mechanism: image and text-only predictions agree \emph{despite} balanced labels, suggesting genuine text-pattern reliance.

To rule out pipeline artifacts in LLaVA-Rad Base's extreme PadChest result, we note that its mean KL divergence between image-conditioned and text-only distributions is $0.26 \pm 0.21$ (nats), with only 0.5\% of samples below $\text{KL} < 0.01$.
The model's softmax distributions do shift when the image is removed; they simply shift within the same binary prediction, consistent with text-reliant behavior rather than a processing artifact.
Moreover, the same model achieves 65.4\% Dangerous on MIMIC (balanced labels), ruling out a PadChest-specific pipeline anomaly.

A per-finding breakdown on Targeted LoRA/PadChest (16 findings with $n \geq 15$) reveals that common findings with high base rates are most Dangerous: vertebral degenerative changes (95.7\%), cardiomegaly (80.4\%), and aortic elongation (85.7\%).
Findings requiring precise localization, such as vascular hilar enlargement (0\%) and infiltrates (0\%), are never Dangerous, suggesting the model must consult the image for spatially specific queries.

\begin{table*}[t]
\centering
\caption{Quadrant distribution across 5 model configurations and 2 datasets. \textbf{Bold} Dangerous fractions exceed 50\%. Models with lower flip rates consistently have higher Dangerous fractions. LLaVA-Rad models have 78 (MIMIC) and 732 (PadChest) evaluable samples.}
\label{tab:quadrant_dist}
\small
\begin{tabular}{ll r rrrr rrrr}
\toprule
Model & Dataset & $n$ & \multicolumn{4}{c}{Count} & \multicolumn{4}{c}{Fraction (\%)} \\
\cmidrule(lr){4-7} \cmidrule(lr){8-11}
 & & & Ideal & Frag. & Dang. & Worst & Ideal & Frag. & Dang. & Worst \\
\midrule
  MedGemma Base & MIMIC & 98 & 31 & 13 & 25 & 29 & 31.6 & 13.3 & 25.5 & 29.6 \\
   & PadChest & 861 & 80 & 412 & 78 & 291 & 9.3 & 47.9 & 9.1 & 33.8 \\
\addlinespace
  Targeted LoRA & MIMIC & 98 & 21 & 13 & 57 & 7 & 21.4 & 13.3 & \textbf{58.2} & 7.1 \\
   & PadChest & 861 & 23 & 220 & 463 & 155 & 2.7 & 25.6 & \textbf{53.8} & 18.0 \\
\addlinespace
  Full LoRA & MIMIC & 98 & 17 & 2 & 77 & 2 & 17.3 & 2.0 & \textbf{78.6} & 2.0 \\
   & PadChest & 861 & 133 & 52 & 523 & 153 & 15.4 & 6.0 & \textbf{60.7} & 17.8 \\
\addlinespace
  LLaVA-Rad Base & MIMIC & 78 & 3 & 13 & 51 & 11 & 3.9 & 16.7 & \textbf{65.4} & 14.1 \\
   & PadChest & 732 & 0 & 5 & 721 & 6 & 0.0 & 0.7 & \textbf{98.5} & 0.8 \\
\addlinespace
  LLaVA-Rad LoRA & MIMIC & 78 & 17 & 8 & 44 & 9 & 21.8 & 10.3 & \textbf{56.4} & 11.5 \\
   & PadChest & 732 & 219 & 38 & 327 & 148 & 29.9 & 5.2 & 44.7 & 20.2 \\
\bottomrule
\end{tabular}
\end{table*}

\subsection{The Consistency--Safety Paradox}
\label{sec:paradox}

\begin{figure}[t]
\centering
\includegraphics[width=\linewidth]{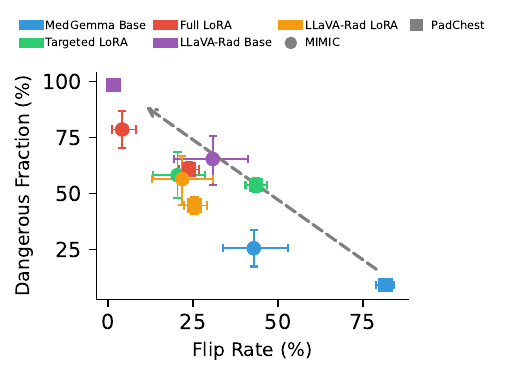}
\caption{Flip rate vs.\ Dangerous fraction with 95\% bootstrap CIs ($n{=}10$ model-dataset combinations; circles: MIMIC; squares: PadChest). The anti-correlation ($r=-0.89$, $\rho=-0.79$) suggests that consistency optimization trades image reliance for apparent reliability. The dashed arrow indicates the direction of increasing danger.}
\label{fig:scatter}
\end{figure}

\cref{tab:quadrant_dist} and \cref{fig:scatter} reveal a negative correlation between flip rate and Dangerous fraction.
Across all $n{=}10$ model-dataset combinations, the Pearson correlation is $r = -0.89$ and the Spearman rank correlation is $\rho = -0.79$, both consistent with a strong monotone relationship, though the small sample size warrants caution in interpreting significance levels.

The extreme case is LLaVA-Rad Base on PadChest: 1.5\% flip rate (the lowest across all settings) yet 98.5\% Dangerous (the highest).
This model would receive the highest possible score on a consistency-only evaluation while being the most dangerous by our taxonomy.
Conversely, MedGemma Base on PadChest has the highest flip rate (81.7\%) but the lowest Dangerous fraction (9.1\%).

This creates what we term the \emph{consistency--safety paradox}: interventions that improve the metric most commonly used to assess reliability (flip rate) systematically degrade the safety property that matters most (image grounding).
The paradox is not limited to a single model or dataset; it emerges consistently across both model families and both evaluation settings.

Within the MedGemma family, the progression is monotonic: Base (42.9\% FR, 25.5\% DF on MIMIC) $\rightarrow$ Targeted LoRA (20.4\% FR, 58.2\% DF) $\rightarrow$ Full LoRA (4.1\% FR, 78.6\% DF).
Each step of consistency improvement corresponds to a step of increased danger.

\subsection{The Accuracy Trap}
\label{sec:accuracy}

\begin{figure}[t]
\centering
\includegraphics[width=\linewidth]{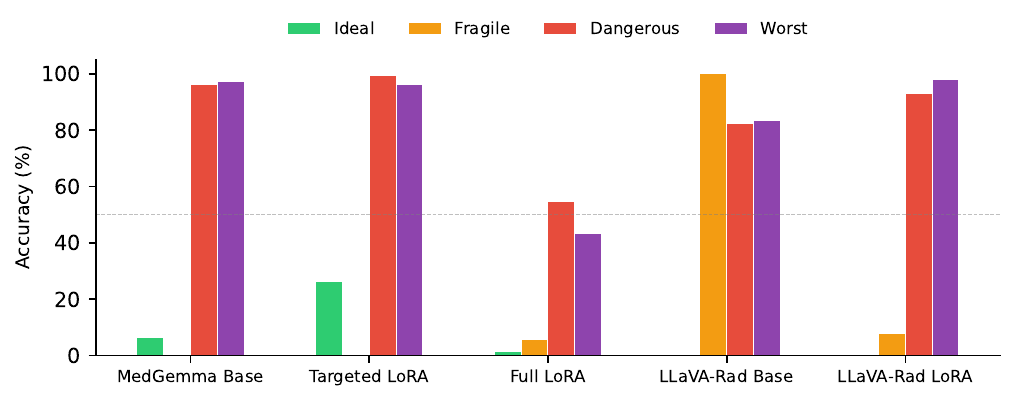}
\caption{Per-quadrant accuracy on PadChest across 5 models. The Dangerous quadrant (red) has higher accuracy than the Ideal quadrant in all models where comparison is possible, creating a paradox where non-image-reliant predictions appear more reliable than image-reliant ones.}
\label{fig:accuracy_bar}
\end{figure}

The Dangerous quadrant is particularly insidious because it often exhibits \emph{higher accuracy} than other quadrants.
\cref{tab:quadrant_accuracy} shows per-quadrant accuracy on PadChest, and \cref{fig:accuracy_bar} provides a visual comparison.

On PadChest, Targeted LoRA achieves 99.6\% accuracy within the Dangerous quadrant vs.\ 26.1\% in Ideal; MedGemma Base shows 96.2\% vs.\ 6.2\%.
In all four models with Ideal samples, Dangerous accuracy exceeds Ideal, often by a large margin.
This occurs because PadChest has 81\% ``yes'' ground truth: a model that consistently predicts ``yes'' without consulting the image is both Dangerous \emph{and} accurate.
The practical implication is severe: accuracy, consistency, and within-subset accuracy all point toward the wrong conclusion.

\begin{table}[t]
\centering
\caption{Per-quadrant accuracy on PadChest (\%). \textbf{Bold}: Dangerous accuracy ${>}80\%$. The Dangerous quadrant has higher accuracy than the Ideal quadrant in all four models where comparison is possible, making it invisible to accuracy-based screening.}
\label{tab:quadrant_accuracy}
\footnotesize
\setlength{\tabcolsep}{3pt}
\begin{tabular}{l rrrr r}
\toprule
Model & Ideal & Fragile & Dangerous & Worst & Overall \\
\midrule
  MedGemma Base & 6.2 & 0.2 & \textbf{96.2} & 97.2 & 42.3 \\
  Targeted LoRA & 26.1 & 0.4 & \textbf{99.6} & 96.1 & 71.7 \\
  Full LoRA & 1.5 & 5.8 & 54.7 & 43.1 & 41.5 \\
  LLaVA-Rad Base & -- & 100.0 & \textbf{82.5} & 83.3 & 82.6 \\
  LLaVA-Rad LoRA & 0.0 & 7.9 & \textbf{93.0} & 98.0 & 61.7 \\
\bottomrule
\end{tabular}
\end{table}

\subsection{Entropy as a Warning Signal}
\label{sec:entropy}

\begin{figure}[t]
\centering
\includegraphics[width=\linewidth]{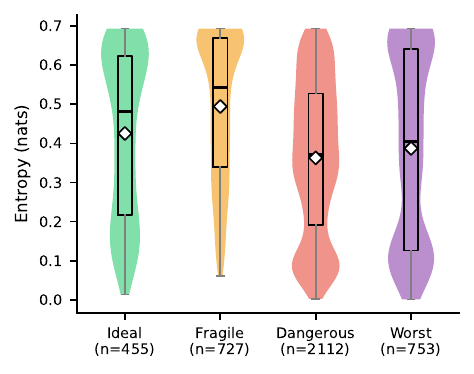}
\caption{Entropy distribution by quadrant across all PadChest models. Dangerous samples have \emph{low} entropy (high confidence), making them invisible to uncertainty-based screening. Fragile samples show higher entropy and are more detectable. Sample counts shown above each violin.}
\label{fig:entropy}
\end{figure}

If Dangerous samples had high predictive entropy, standard uncertainty-based screening could flag them at deployment time.
\cref{fig:entropy} shows that this is not the case.
Across all PadChest models combined, the Dangerous quadrant has \emph{lower} mean entropy than the Fragile quadrant.
Dangerous samples are not only consistent and accurate; they are also highly confident.

The explanation is straightforward: a model applying a text-pattern shortcut does so with high confidence because the heuristic is reliable.
The Fragile quadrant, in contrast, shows elevated entropy and is partially detectable through uncertainty estimation.
This asymmetry means an uncertainty-based deployment gate would filter out image-reliant samples (removing Fragile) while passing samples whose predictions are unchanged by image removal (Dangerous).

\cref{tab:qualitative} illustrates one concrete sample per quadrant from Targeted LoRA on PadChest.
The Ideal sample (NSG tube) correctly predicts ``yes'' with image and ``no'' without, confirming genuine visual reasoning.
The Dangerous sample (acute abnormality) produces the same ``no'' prediction regardless of image presence, with similar softmax probabilities, exemplifying the false-reliability failure mode.

\begin{table}[t]
\centering
\caption{Qualitative examples: one sample per quadrant (Targeted LoRA, PadChest). Img/Text columns show predictions with/without image. Dangerous samples have near-identical $p_\text{yes}$ values.}
\label{tab:qualitative}
\footnotesize
\setlength{\tabcolsep}{2pt}
\begin{tabular}{l l cc cc c}
\toprule
Quadrant & Question & Img & Text & $p_{\text{yes}}^{\text{img}}$ & $p_{\text{yes}}^{\text{txt}}$ & GT \\
\midrule
  Ideal & Is there NSG tube? & yes & no & 0.82 & 0.41 & yes \\
  Fragile & Is there infiltrates? & no & yes & 0.07 & 0.79 & yes \\
  Dangerous & Is there any acute abnormality on thi... & no & no & 0.07 & 0.18 & no \\
  Worst & Is there NSG tube? & no & no & 0.38 & 0.41 & yes \\
\bottomrule
\end{tabular}
\end{table}

\section{Discussion}
\label{sec:discussion}

\paragraph{The deployment trap.}
Our results expose a critical gap in current medical VLM evaluation practice.
A deployment pipeline that checks accuracy, flip rate, and predictive confidence would \emph{preferentially select} the most dangerous models.
Consider LLaVA-Rad Base on PadChest: it passes all three checks: 82.5\% accuracy (above random baseline), 1.5\% flip rate (lowest of all models), and high mean confidence (low entropy). Yet 98.5\% of its predictions are unchanged when the image is removed.
A clinician relying on this model would receive confident, consistent, and usually correct answers that are decision-invariant to the specific radiograph under our binary test (\cref{eq:image_reliant}).

This is not a corner case.
Across our evaluation, 9 of 10 model-dataset combinations have Dangerous fractions exceeding 25\%, and 7 of 10 exceed 50\%.
The consistency--safety paradox means that efforts to improve deployment readiness through consistency training can \emph{worsen} the underlying safety failure.

\paragraph{The LoRA paradox.}
LoRA fine-tuning for consistency shifts samples from Fragile into Dangerous across both model families: on PadChest, Dangerous increases monotonically from 9.1\% (Base) to 53.8\% (Targeted LoRA) to 60.7\% (Full LoRA).
The mechanism is intuitive: the easiest way to eliminate paraphrase sensitivity is to learn text shortcuts that make predictions invariant to image removal.
Full LoRA, with access to all 34 layers, has the greatest capacity for such shortcuts.
LLaVA-Rad LoRA partially breaks this pattern (44.7\% Dangerous vs.\ 98.5\% for Base, with the highest Ideal fraction 29.9\%), suggesting that the outcome is model-dependent.

\paragraph{Practical recommendation.}
We recommend that any deployment evaluation of medical VLMs include a \textbf{text-only baseline test}: running each evaluation sample through the model with the image removed.
This requires exactly one additional forward pass per sample, adding negligible computational overhead, and immediately reveals whether consistency is grounded in visual evidence.
A sample that produces the same prediction with and without the image should be flagged regardless of its accuracy or consistency.

The four-quadrant framework provides a complete classification: rather than reporting a single flip rate, evaluations should report the full quadrant distribution.
This enables stakeholders to distinguish between models that are reliable for the right reasons (high Ideal fraction) versus those that merely appear reliable (high Dangerous fraction).

We propose a five-step deployment checklist: compute flip rates, run a text-only baseline, classify samples into the four quadrants, report the full distribution alongside aggregate metrics, and flag any model with Dangerous fraction ${>}50\%$.

\paragraph{Complementary grounding checks.}
We validated \cref{eq:image_reliant} with two independent experiments.
\textbf{KL divergence} between image-conditioned and text-only distributions achieves AUROC${}=0.76$ for detecting Dangerous samples (mean $\text{KL}$: Dangerous${}=0.12$ vs.\ Ideal${}=1.14$).
An \textbf{image-swap test} across all 4{,}497 samples (5 random replacement images per sample) shows Dangerous samples are 78--97\% swap-invariant vs.\ 26--77\% for Ideal, and a \textbf{null-image} test (black 224${\times}$224) confirms 97--100\% of Dangerous samples produce identical predictions without any visual input.

\paragraph{Implications for consistency-based training.}
Our findings do not imply that consistency training is harmful: paraphrase sensitivity is a genuine failure mode that can cause real clinical harm when semantically equivalent questions yield contradictory answers.
However, consistency objectives should be paired with image-grounding constraints.
A training loss that rewards consistency \emph{only when the prediction differs from the text-only baseline} would avoid the Dangerous quadrant by design, encouraging the model to achieve consistency through improved visual reasoning rather than text shortcuts.
We leave the design and evaluation of such grounding-aware consistency objectives to future work.

\paragraph{Limitations.}
Our evaluation covers binary (yes/no) questions on two chest X-ray datasets; quadrant distributions may differ for open-ended generation, multi-class tasks, or other modalities.
The text-only baseline is a coarse measure: a model might use the image to calibrate confidence without changing the final prediction, which our binary test would not capture (though our image-swap and null-image experiments corroborate the binary findings).
Dataset sizes are limited (MIMIC: 98; PadChest: 861), PadChest label imbalance (81\% ``yes'') amplifies affirmative-bias Dangerous rates, and our five configurations span only two model families.
A theoretical analysis of when consistency training leads to text shortcuts would strengthen the framework.

\paragraph{Future directions.}
The text-only baseline could be augmented with gradient-based attribution~\cite{selvaraju2017gradcam} for a continuous image-reliance measure, and applied longitudinally during fine-tuning to identify when text shortcuts emerge.
Extending to open-ended generation would require semantic similarity for consistency and counterfactual image editing for reliance.
A grounding-aware consistency loss that rewards consistency only when the prediction differs from the text-only baseline could avoid the Dangerous quadrant by design.

\paragraph{Broader impact.}
Current evaluation practice for medical AI often emphasizes aggregate performance metrics such as accuracy and sensitivity~\cite{rajpurkar2022ai} and may miss modality-grounding failures of the kind we identify.
Our results show that this gap can allow systems whose predictions are decision-invariant to image removal to pass standard evaluation.
Modality-grounding tests such as text-only baseline comparison add negligible cost (one forward pass) and should be considered as part of deployment evaluation for medical VLMs.

\section{Conclusion}
\label{sec:conclusion}

We introduced a four-quadrant per-sample safety taxonomy that jointly evaluates consistency and image reliance in medical VLMs.
Our evaluation across five model configurations and two chest X-ray datasets reveals a consistency--safety paradox: models with the lowest flip rates have the highest fraction of Dangerous samples: predictions that are consistent, accurate, and confident, yet unchanged when the image is removed.

This finding has immediate practical implications: flip rate alone is an insufficient and potentially misleading metric for deployment clearance.
We recommend always pairing consistency evaluation with a text-only baseline to expose the false reliability trap.
The four-quadrant framework provides a simple, actionable decomposition that requires no architectural changes, only one additional forward pass per evaluation sample.

The broader lesson is that evaluation metrics in medical AI must be interpreted jointly rather than in isolation.
Consistency is necessary but not sufficient for safety.
Accuracy is necessary but can be achieved through shortcuts.
Confidence is desirable but can arise from reliable heuristics rather than correct reasoning.
Only by examining these properties together, at the per-sample level, can we distinguish models that are genuinely safe from those that merely appear so.

{
    \small
    \bibliographystyle{ieeenat_fullname}
    \bibliography{references}
}

\end{document}